%% file: KDD.tex
\documentclass[sigconf]{acmart}

\usepackage{algorithm}
\usepackage{algorithmic}
\usepackage{amsthm}
\theoremstyle{plain}
\usepackage{xcolor} 
\usepackage[table]{xcolor}
\definecolor{MyRowColor}{HTML}{FDF0E7}
\usepackage{tikz}
\usepackage{amsmath}   
\usepackage{colortbl}
\usepackage[dvipsnames,table,svgnames]{xcolor}

\usepackage{booktabs}
\usepackage{multirow}
\usepackage{enumitem} 

\usepackage{newfloat}
\usepackage{listings}
\DeclareCaptionStyle{ruled}{labelfont=normalfont,labelsep=colon,strut=off} 
\floatstyle{ruled}
\newfloat{listing}{tb}{lst}{}
\floatname{listing}{Listing}
\AtBeginDocument{%
  }

\copyrightyear{2026}
\acmYear{2026}
\setcopyright{cc}
\setcctype{by}
\acmConference[KDD 2026] {Proceedings of the 32nd ACM SIGKDD Conference on Knowledge Discovery and Data Mining V.2}{August 9--13, 2026}{Jeju Island, Republic of Korea.}
\acmBooktitle{Proceedings of the 32nd ACM SIGKDD Conference on Knowledge Discovery and Data Mining V.2 (KDD 2026), August 9--13, 2026, Jeju Island, Republic of Korea}
\acmISBN{979-8-4007-2259-2/2026/08}
\acmDOI{10.1145/3770855.3817801}

\begin{document}

\title{TRACE: Discovering Task-Specific Parameter via Adaptation-Aware Probing for Continual Fine-Tuning}

\author{Xiaosong Han}
\authornote{Key Laboratory of Symbolic Computation and Knowledge Engineering of the Ministry of Education}
\affiliation{
  \institution{College of Computer Science and Technology, Jilin University}
  \city{Changchun}
  \country{China}
}
\email{hanxiaosong@jlu.edu.cn}

\author{Ke Chen}
\authornotemark[1]
\affiliation{
  \institution{College of Software, Jilin University}
  \city{Changchun}
  \country{China}
}
\email{chenke24@mails.jlu.edu.cn}

\author{Xindi Dai}
\authornotemark[1]
\affiliation{
  \institution{College of Computer Science and Technology, Jilin University}
  \city{Changchun}
  \country{China}
}
\email{daixd24@mails.jlu.edu.cn}

\author{Di Liang}
\author{Minlong Peng}
\affiliation{
  \institution{Fudan University}
  \city{Shanghai}
  \country{China}
}
\email{{liangd17,mlpeng16}@fudan.edu.cn}

\author{Wei Pang}
\affiliation{
  \institution{School of Mathematical and Computer Sciences, Heriot-Watt University}
  \city{Edinburgh}
  \country{Scotland}
}
\email{w.pang@hw.ac.uk}

\author{Fausto Giunchiglia}
\affiliation{%
  \institution{Department of Information Engineering and Computer Science, University of Trento}
  \city{Trento}
  \country{Italy}
}
\email{fausto.giunchiglia@unitn.it}

\author{Xiaoyue Feng}
\authornotemark[1]
\authornote{Corresponding author}
\affiliation{%
  \institution{College of Computer Science and Technology, Jilin University}
  \city{Changchun}
  \country{China}
}
\email{fengxy@jlu.edu.cn}

\author{Yonghao Liu}
\authornotemark[1]
\authornotemark[2]
\affiliation{
  \institution{College of Computer Science and Technology, Jilin University}
  \city{Changchun}
  \country{China}
}
\email{yonghao20@mails.jlu.edu.cn}

\author{Renchu Guan}
\authornotemark[1]
\authornotemark[2]
\affiliation{%
  \institution{College of Computer Science and Technology, Jilin University}
  \city{Changchun}
  \country{China}
}
\email{guanrenchu@jlu.edu.cn}

\renewcommand{\shortauthors}{Xiaosong Han et al.}

\begin{abstract}
In real-world deployment, LLMs are often adapted \emph{continually} across tasks to keep LLMs up-to-date in production, where new fine-tuning should preserve previously learned skills.
However, indiscriminately mixing tasks can dilute task specialization, while sequential fine-tuning (full-parameter or low rank adaptation) often causes catastrophic forgetting due to destructive overwriting.
Replay-based continual tuning and maintaining separate task-specific adapters can mitigate forgetting, but introduce additional compute, storage, and management overhead.
Recognizing the redundancy of LLM parameters for any single task, we reframe continual task adaptation as \emph{task-specific parameter discovery} via adaptation-aware probing:
a short warm-start probe exposes a task's adaptation trace, enabling us to identify and isolate the small subset of parameters essential for each task to mitigate catastrophic forgetting. 
Building on this view, we introduce \textbf{TRACE}, a novel approach for discovering \textbf{T}ask-specific pa\textbf{R}ameters via \textbf{A}daptation-aware probing for \textbf{C}ontinual fin\textbf{E}-tuning.
We perform a short warm-start fine-tune to derive task-specific core parameters by comparing the warm-started and pre-trained models.
Core parameters are identified via two strategies: importance scoring (L$_2$ norm and Fisher Information) and specificity analysis (cosine similarity of parameter updates).
In continual fine-tuning settings, only the active task’s core parameters are updated while others remain frozen, preserving prior knowledge.
We conduct extensive experiments across multiple standard benchmarks to demonstrate the superior performance of our proposed method.
Additionally, we validate the generalization of our method through a cross-model and scale transferability study, demonstrating a ``small-to-large'' paradigm that guides the fine-tuning of large-scale models under resource constraints.
Our public code can be found \textcolor{red}{\href{https://github.com/KEAML-JLU/TRACE}{here}}.
\end{abstract}

\ccsdesc[500]{Computing methodologies~Natural language processing}
\ccsdesc[500]{Computing methodologies~Knowledge representation and reasoning}
\ccsdesc[300]{Computing methodologies~Natural language generation}

\keywords{Large Language Model, Continual Fine-Tuning}

\maketitle

\input{introduction}
\input{related_work}
\input{method}
\input{experiment}
\input{result}
\input{conclusion}
\begin{acks}
Our work is supported by the National Natural Science Foundation of China (No. 62372209 and No. 62372494). Fausto Giunchiglia's work is funded by European Union's Horizon 2020 FET Proactive Project (No. 823783).
\end{acks}
\bibliographystyle{ACM-Reference-Format}
\balance
\bibliography{kdd2026}

\appendix
\renewcommand{\thetable}{A.\arabic{table}}
\captionsetup[table]{labelformat=simple, labelsep=colon, name=Table}
\setcounter{table}{0}

\renewcommand{\thefigure}{A.\arabic{figure}}
\captionsetup[figure]{labelformat=simple, labelsep=colon, name=Figure}
\setcounter{figure}{0}

\input{appendix}

\end{document}

%% file: introduction.tex
\section{Introduction}
Large Language Models (LLMs) have demonstrated impressive generalization capabilities across a wide range of tasks \cite{ouyang2022training, liu2021deep, li2026simple, liu2025boosting, liu2025high}. A key enabler of this versatility is Supervised Fine-Tuning (SFT), which allows general-purpose models to be specialized for high-stakes tasks such as mathematical reasoning, medical question answering, and complex programming tasks \cite{zhao2023survey,lu2022makes,liu2023local}. This capability is crucial for real-world applications, where fine-tuning transforms a general-purpose model into a competent task-specific assistant. However, most existing SFT approaches focus on a single-task setting, neglecting the practical need for continual fine-tuning, where tasks arrive over time in production. In contrast, proprietary models such as GPT-5 \cite{openai2025gpt5} demonstrate remarkable versatility, suggesting that public LLMs need to learn to handle multiple specialized tasks over time. However, the conventional SFT paradigm harbors a fundamental flaw: it inherently operates as a destructive update mechanism. When an LLM is continually fine-tuned on multiple tasks, new knowledge often comes at the expense of previously acquired skills. 
\begin{figure}[ht]
    \centering \includegraphics[width=0.46\textwidth]{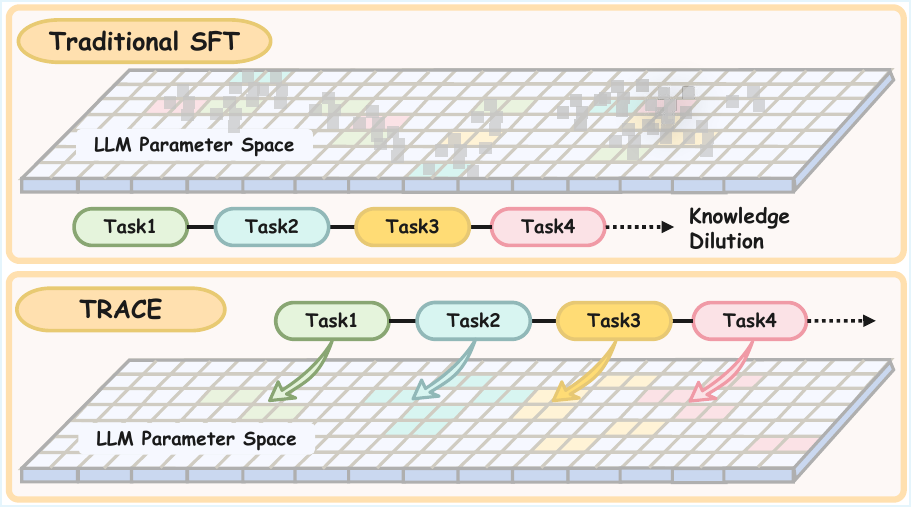}
    \caption{A parameter-space view of continual fine-tuning. Traditional SFT dilutes task knowledge over time, while TRACE discovers task-specific core parameters and selectively activates them to preserve prior skills.}
    \label{intro}
\end{figure}
\noindent
For example, fine-tuning a model initially trained on code generation with medical question answering data may overwrite its internal representations of programming logic. The resulting model often exhibits degraded performance on earlier tasks, effectively becoming a hybrid expert that is neither proficient in code nor accurate in medicine. This phenomenon, known as catastrophic forgetting \cite{luo2023empirical,wang2023trace,de2021continual}, underscores a core limitation of current fine-tuning strategies: the absence of mechanisms for isolating task-specific knowledge.

Existing approaches to this problem—such as experience replay~\cite{DMT}, which incurs substantial computational and memory overhead; parameter-efficient adapter-based adaptation\cite{hu2022lora}, which reduces the number of trainable parameters per task but still faces trade-offs between forgetting (when reusing a shared adapter) and storage complexity (when maintaining task-specific adapters); mixing all tasks into a joint SFT process~\cite{albalaksurvey}, which is often impractical in continual fine-tuning since it requires persistent access to historical data and may still suffer from task interference; and regularization-based methods~\cite{lin2023speciality,smith2023continual}, which treat the model as a monolithic entity and impose global constraints to mitigate forgetting. However, they fail to address the root cause of the issue: the unregulated overwriting of task-specific internal parameters in continual fine-tuning settings. In essence, these methods remain committed to the destructive paradigm, offering only incremental safeguards without explicitly identifying and isolating the parameters responsible for new-task adaptation. Consequently, a natural question arises: \textit{Can we design a task-specific fine-tuning mechanism that enables an LLM to acquire new capabilities while preserving its previously learned knowledge?} As illustrated in Fig. \ref{intro}, conventional sequential SFT broadly updates shared parameters, progressively diluting earlier task knowledge and causing catastrophic forgetting. Our method instead identifies task-specific core parameters and selectively updates them, enabling continual adaptation without losing previously learned capabilities. 

In neuroscience, it is well established that different regions of the brain are specialized for distinct functions \cite{johnson2001functional}---for example, the temporal lobe is responsible for processing auditory signals, whereas the occipital lobe is involved in perceiving and processing basic visual features. Drawing an analogy to this mechanism, prior studies have made an important observation: different layers and modules within LLMs contribute unequally across tasks, with certain layers playing dominant roles in specific domains while others remain relatively inactive \cite{BansalGDBKR23,ding2023parameter,GevaSBL21}. This structural specialization suggests a promising direction---selectively tuning task-relevant layers while freezing others preserves prior knowledge and reduces interference. Motivated by this insight, we reframe continual fine-tuning as \emph{task-specific parameter discovery via adaptation-aware probing}. Accordingly, we introduce TRACE, a novel framework that enables LLMs to acquire new capabilities while preserving previously learned knowledge with low training overhead, TRACE performs the following:

\begin{itemize}[leftmargin=1.8em,itemsep=0.5em,topsep=0.5em]
    \item \textbf{Task-Specific Core Parameter Discovery}: For each task in the continual stream, 
    TRACE performs \emph{adaptation-aware probing} via a brief warm-start fine-tune, and compute the parameter update vectors. To identify task-specific core parameters, we adopt two complementary strategies: a magnitude-based approach that combines L$_2$  norm changes and Fisher Information to evaluate parameter importance, and a direction-based approach that analyzes the cosine similarity of update vectors across tasks to capture task-specificity.

    \item \textbf{Selective Parameter Activation}: During continual fine-tuning, we only activate the core parameters relevant to the current task while keeping all others frozen. This protects prior knowledge and minimizes destructive updates.
\end{itemize}
 Futhermore, we perform a cross-model transferability study, showing that core parameters identified by TRACE on a small model (LLaMA3-8B) can be transferred to guide the fine-tuning of larger, architecturally different models (\textit{\textit{e.g.}}, Qwen2.5-14B). This validates that TRACE uncovers a task-centric ``functional blueprint'' rather than a model-specific artifact, and offers a ``small-to-large'' transfer paradigm, where knowledge from a computationally inexpensive 8B model is successfully mapped to guide the fine-tuning of 14B models. Finally, extensive experiments across various foundational LLMs show that TRACE consistently outperforms sequential full fine-tuning and other baselines on downstream tasks.

%% file: related_work.tex
\section{Related Work}
\label{related_work}
\subsection{Catastrophic Forgetting in SFT of LLMs}
SFT is a key step in adapting LLMs to downstream tasks, where the model is optimized on curated instruction-response pairs to enhance its task-specific performance \cite{DMT,pareja2024unveiling}. While single-task SFT has achieved impressive results in specialized domains, real-world applications increasingly demand multi-task SFT---adapting LLMs to multiple domains such as mathematical reasoning, code generation, and medical question answering. This setting introduces a fundamental challenge: models fine-tuned on one task often forget previously acquired knowledge when adapted to new tasks, a phenomenon known as catastrophic forgetting \cite{luo2023empirical,lin2023speciality}. Prior work has explored various strategies to alleviate catastrophic forgetting. Regularization-based methods constrain parameter drift during fine-tuning but fail to account for task-specific relevance\cite{panigrahi2023task}. Weight-based approaches (\textit{e.g.}, Wise-FT \cite{wortsman2022robust}, V-SoftMask \cite{ke2023continual}) interpolate between model states or scale gradients by importance scores, yet often lack sufficient granularity to avoid inter-task interference. Architecture-based methods such as LoRA \cite{hu2022lora} or adapters freeze the backbone and introduce task-specific modules to isolate knowledge, but increase memory cost and may underutilize shared capacity. 
Recent studies point out that LLMs contain substantial parameter redundancy, and different tasks tend to activate distinct functional subspaces \cite{GevaSBL21,meng2022locating}. This suggests that preserving task performance does not require tuning all parameters; instead, updating a small task-specific subset may suffice. Inspired by this, our work investigates core parameter identification and selective activation to mitigate forgetting in multi-task SFT.

\subsection{Insights into Transformer Module Specialization}
Recent interpretability studies reveal that Transformer-based LLMs exhibit significant modular specialization: different components contribute unequally across tasks, and many parameters are underutilized. For instance, feed-forward networks and attention heads display strong task-dependent activation patterns, with certain heads or neurons being consistently active only under specific task conditions \cite{michel2019sixteen,GevaSBL21}. Complementary empirical analyses demonstrate widespread redundancy in deep models: over 80\% of neurons in pretrained LLMs can often be pruned with minimal impact on performance, suggesting that full-parameter tuning is both computationally inefficient and prone to unnecessary interference \cite{sun2023simple,dalvi2020analyzing}. Similarly, layer-wise Shapley value analysis reveals that certain ``cornerstone'' layers dominate error reduction, while omitting others has negligible effects---highlighting the uneven importance distribution across layers \cite{zhang2024investigating}. These insights have motivated techniques for selective tuning of high-impact components, often guided by empirical analysis \cite{meng2022locating,allen2023physics}.
Building on these insights, our work proposes a task-specific core parameter identification strategy tailored for multi-task SFT. By isolating and preserving the task-specific parameters learned in earlier stages, we enable subsequent tasks to fine-tune only their own relevant subspaces. 

%% file: method.tex
\section{Method}
\begin{figure*}[ht]
    \centering
    \includegraphics[width=0.95\textwidth]{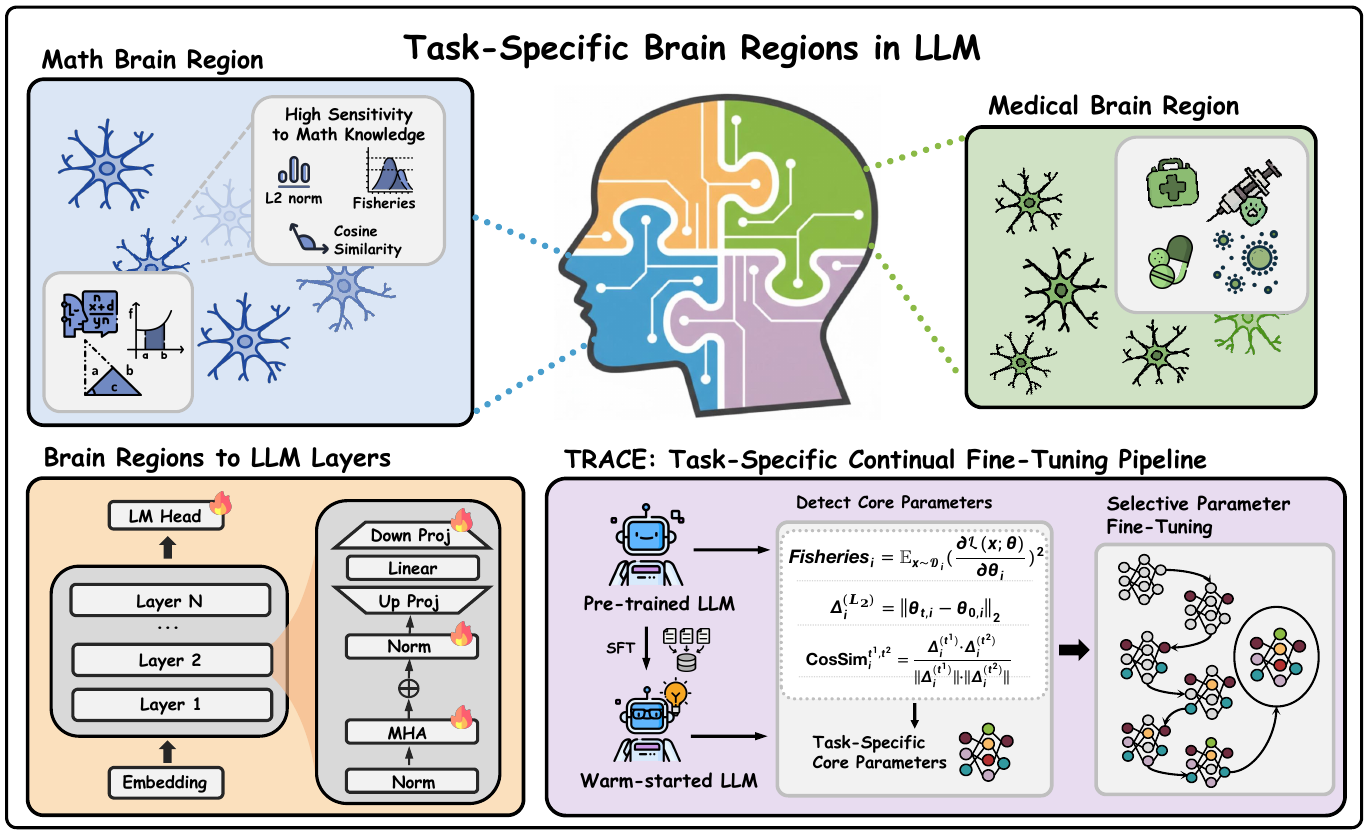}
    \caption{The framework of TRACE. The top part illustrates the core intuition behind TRACE: we conceptualize the LLM as a brain, where different regions are activated by different task-specific tasks. The bottom left visualizes how this brain-like activation mechanism maps to the LLM’s Transformer architecture. The bottom right illustrates the TRACE pipeline: a pre-trained LLM is first fine-tuned on task-specific data. Core parameters are then identified by comparing the original and fine-tuned models using L$_2$  distance, Fisher information, and cosine similarity. These parameters are selectively activated for continual fine-tuning, while others are frozen to retain prior knowledge.}
    \label{framework}
\end{figure*}

We formalize catastrophic forgetting in continual SFT and then present TRACE, which consists of \textit{task-specific core parameter discovery} and \textit{selective parameter activation}. An overview of the workflow is shown in Fig. \ref{framework}.

\subsection{Problem Definition}
LLMs aligned with human preferences exhibit strong general capabilities, but task-specific applications (\textit{e.g.}, code, math) typically still require fine-tuning on supervised datasets. In continual fine-tuning, adapting to a new task often degrades performance on earlier tasks, \textit{i.e.}, catastrophic forgetting.

Formally, let $\theta_0$ denote the parameters of the initial aligned LLM. Given a sequence of $\mathit{K}$ tasks $T = \{T_1, T_2, \ldots, T_K\}$, the goal of continual SFT is to update the model parameters through multiple stages of task-specific training: 
\begin{equation}
\theta_k = \arg\min_{\theta} \mathcal{L}_{\text{SFT}}(\theta; T_k), \quad \text{with} \quad \theta \leftarrow \theta_{k-1},
\end{equation}
where $\mathcal{L}_{\text{SFT}}(\theta; T_k)$ denotes the fine-tuning loss on task $T_k$, and $\theta_k$ represents the model parameters after training on the $\mathit{k}$-th task. The objective is to obtain a single model $\theta_K$ that performs well on all tasks $\{T_1, T_2, \ldots, T_K\}$. However, full-parameter fine-tuning often leads to a significant drop in performance on previously learned tasks:
\begin{equation}
\text{Performance}(\theta_k; T_{<k}) \ll \text{Performance}(\theta_{k-1}; T_{<k}).
\end{equation}
This inequality reflects the destructive nature of standard sequential fine-tuning, where learning a new task often comes at the cost of forgetting earlier ones.

To address this challenge, we leverage the architectural characteristics of Transformer-based LLMs (\textit{e.g.}, LLaMA \cite{touvron2023llama}, Qwen \cite{qwen2.5}), which consist of a stack of decoder layers, each composed of a multi-head attention module and a feed-forward network. Prior work suggests that different modules and layers contribute differently to task specialization and generalization. This motivates us to identify and selectively update task-relevant parameters in order to preserve knowledge from earlier tasks.

\subsection{Discovering Task-Specific Parameter and Selective Parameter Activation}
We now describe the two main components of TRACE.

\noindent\textbf{\textit{Step 1: Task-Specific Core Parameter Discovery.}}
We propose two independent strategies for identifying task-specific core parameters. Both methods aim to select a subset of parameters that are most crucial for a given task, but they do so from distinct perspectives: one based on magnitude and sensitivity, and the other based on task-specificity in update direction.

    \noindent\textbullet\ \textbf{\textbf{L$_2$-Fisher-Based Importance Scoring}}: The first strategy evaluates how significantly each parameter contributes to a task, by combining the magnitude of change and the sensitivity of the loss function. Given the base model parameters $\theta_0$  and the task-specific fine-tuned parameters $\theta_k^*$. It is crucial to note that instead of a costly multi-epoch training process, our approach only requires a short warm-start fine-tune for each task (a single epoch). The objective of this step is not to achieve optimal task performance, but to efficiently generate a strong signal indicating which parameters are most important and specific to a given domain. We compute the L$_2$  norm difference of parameter $\theta^{(i)}$ as:
    \begin{equation}
    \text{L$_2$}^{(i)} = \|\theta_k^{*(i)} - \theta_0^{(i)}\|_2.
    \end{equation}
    This reflects how much the parameter changed during task-specific tuning. To further assess importance, we compute the Fisher information of each parameter:
    \begin{equation}
    \text{Fisher}^{(i)} = \mathbb{E}_{x \sim \mathcal{D}_k} \left[ \left( \frac{\partial \mathcal{L}_{\text{SFT}}(x; \theta_k^{(i)})}{\partial \theta^{(i)}} \right)^2 \right].
    \end{equation}
    This captures how sensitive the task loss is to perturbations in parameter $\theta^{(i)}$. We normalize both metrics and compute a combined score. We set $\alpha=\beta=0.5$ by default, which is validated by the sensitivity analysis in Sec.~\ref{sec:alpha_sens}:
    \begin{equation}
    \text{score}^{(i)} = \alpha \cdot \overline{\text{L$_2$ }}^{(i)} + \beta \cdot \overline{\text{Fisher}}^{(i)}.
    \end{equation}
    Parameters are then ranked by $\text{score}^{(i)}$, and the top $r\%$ are selected as the task's core parameters; the choice of $r$ is analyzed in Sec.~\ref{sec:r_sens}.
    
    \noindent\textbullet\ \textbf{\textbf{Cosine Similarity-Based Specificity Scoring}}: The second strategy identifies core parameters based on how uniquely each parameter is updated for a given task compared to others. The intuition is that if a parameter’s update direction is very different from that of other tasks, then it likely encodes task-specific knowledge. We first compute the parameter difference vector:
    \begin{equation}
        \Delta_k^{(i)} = \theta_k^{*(i)} - \theta_0^{(i)}.
    \end{equation}
    Then, for each parameter and pair of tasks $(T_k, T_j)$, we compute the cosine similarity of their updates:
    \begin{equation}
        \text{Sim}_{k,j}^{(i)} = \frac{\langle \Delta_k^{(i)}, \Delta_j^{(i)} \rangle}{\|\Delta_k^{(i)}\|_2 \cdot \|\Delta_j^{(i)}\|_2}.
    \end{equation}
    Finally, we define a specificity score for task $T_k$ as:
    \begin{equation}
        \text{Spec}_k^{(i)} = 1 - \frac{1}{K-1} \sum_{\substack{j=1, \; j \neq k}}^K \text{Sim}_{k,j}^{(i)}.
    \end{equation}
    A high specificity score indicates that the parameter is adjusted in a task-unique way. We rank parameters by $ \text{Spec}_k^{(i)}$ and select the top $r\%$ as the core parameters set $\mathcal{C}_k$ for task $T_k$.

\noindent\textbf{\textit{Step 2: Selective Parameter Activation}.} After identifying task-specific core parameters for each task, we perform continual fine-tuning in a way that only updates the parameters most relevant to the current task, while minimizing interference with parameters important to previous tasks.

For each task $T_k$, we define a core parameter set $\mathcal{C}_k$, selected using one of the two strategies described in Step 1. During training on $T_k$, only parameters in $\mathcal{C}_k$ are activated and allowed to update, while all other parameters remain frozen. Importantly, we allow parameter reuse across tasks. That is, if a parameter is shared between multiple tasks’ core sets, it can be updated again when it reappears in the current task's core set. We do not enforce strict freezing of all previous core parameters, only those that are irrelevant to the current task. Formally, the parameter update rule at step $\mathit{k}$ becomes:
\begin{equation}
    \theta_k \leftarrow \theta_{k-1} - \eta \cdot \nabla_{\theta^{(i)} \in \mathcal{C}_k} \mathcal{L}_{\text{SFT}}(\theta; T_k),
\end{equation}
where $\eta$ is the learning rate, and $\mathcal{C}_k$ denotes the task-specific core parameter set. Parameters not in $\mathcal{C}_k$ (\textit{i.e.}, $\theta^{(i)} \notin \mathcal{C}_k$) are excluded from gradient updates. 

This selective activation mechanism ensures that only task-relevant parameters are adjusted at each stage, preserving previously learned knowledge and minimizing destructive interference across tasks. To facilitate the understanding, we detail the specific training process of our method in Algorithm \ref{TRACE}. 

\begin{algorithm}
\caption{TRACE: Discovering Task-Specific Parameter and Selective Parameter Activation}
\label{TRACE}
\begin{algorithmic}[1]
\REQUIRE Initial LLM parameters $\theta_0$, sequence of tasks $T = \{T_1, T_2, \dots, T_K\}$, core parameter ratio $r$
\ENSURE Final model parameters $\theta_K$

\STATE \textcolor{gray}{\# Step 1: Core Parameter Identification (Performed Offline)}
\STATE \textcolor{gray}{\# First, generate all warm-started models and parameter deltas}
\FOR{each task $T_k \in T$}
    \STATE Perform a short warm-start fine-tune: $\theta_k^* \leftarrow \texttt{WarmStart}(\theta_0, T_k)$
    \STATE Store parameter deltas: $\Delta_k \leftarrow \theta_k^* - \theta_0$
\ENDFOR

\STATE \textcolor{gray}{\# Then, select core parameters using one of the two strategies below}
\STATE \textbf{Strategy A: L$_2$-Fisher Importance Scoring}
\FOR{each task $T_k \in T$}
    \STATE Compute L$_2$ norm of deltas: L$_{2,k} \leftarrow ||\Delta_k||_2$
    \STATE Compute Fisher Information on task $T_k$: $\text{Fisher}_k \leftarrow \texttt{ComputeFisher}(\theta_0, T_k)$
    \STATE Combine scores: $\mathit{Score}_k \leftarrow \alpha \cdot \overline{\text{L}_{2,k}} + \beta \cdot \overline{\text{Fisher}_k}$
    \STATE Select top-$r\%$ core parameters $\mathcal{C}_k$ based on $\mathit{Score}_k$
\ENDFOR

\STATE \textbf{Strategy B: Cosine Similarity Specificity Scoring}
\FOR{each task $T_k \in T$}
    \STATE Compute average similarity with all other tasks: \\
    \hspace{1em} $AvgSim_k \leftarrow \frac{1}{K-1} \sum_{j \neq k} \texttt{CosineSimilarity}(\Delta_k, \Delta_j)$
    \STATE Compute specificity score: $Spec_k \leftarrow 1 - AvgSim_k$
    \STATE Select top-$r\%$ core parameters $\mathcal{C}_k$ based on $Spec_k$
\ENDFOR

\STATE \textcolor{gray}{\# Step 2: Sequential Selective Fine-tuning}
\STATE Initialize $\theta \leftarrow \theta_0$
\FOR{$k = 1$ to $K$}
    \STATE Activate parameters in the chosen core set $\mathcal{C}_k$ and freeze all others
    \STATE Fine-tune on $T_k$ with gradients only on $\mathcal{C}_k$: \\
    \hspace{1em} $\theta \leftarrow \theta - \eta \cdot \nabla_{\theta^{(i)} \in \mathcal{C}_k} \mathcal{L}_{\text{SFT}}(\theta; T_k)$
\ENDFOR
\STATE \textbf{return} $\theta_K \leftarrow \theta$
\end{algorithmic}
\end{algorithm}

%% file: experiment.tex
\section{Experiments}
\subsection{SFT Datasets}
We consider three task-specific SFT datasets: \textbf{Code Alpaca} \cite{codealpaca} for code generation, \textbf{GSM8K-RFT} \cite{yuan2023scaling} for mathematical reasoning, and \textbf{MedQA} \cite{jin2021disease} for medical question answering. These tasks represent distinct domains with potentially conflicting optimization directions. Detailed descriptions of these SFT datasets are available in \textbf{Appendix} \ref{detail_sft}.

\subsection{Foundation Models}
To validate the generality of our approach, We evaluate TRACE on diverse open-source LLM backbones and scales, including \textbf{LLaMA2} (7B)\cite{touvron2023llama}, \textbf{LLaMA3} (8B)~\cite{dubey2024llama}, \textbf{Qwen2.5} (14B)~\cite{qwen2.5}, \textbf{Qwen3} (32B)~\cite{yang2025qwen3}, and the distilled \textbf{DeepSeek-R1} (1.5B)~\cite{guo2025deepseek}. 
This diverse selection shows that TRACE's gains generalize beyond any single backbone. Model details are in \textbf{Appendix}~\ref{detail_model}.

\subsection{Baselines}
To validate the benefits of TRACE, we compare it with a diverse set of representative fine-tuning strategies that reflect common approaches to continual adaptation. \textbf{Joint Fine-tuning} \cite{zhan2024towards} directly mixes all SFT data for joint fine-tuning, while the \textbf{Sequential Fine-tuning} \cite{hu2025fine} applies sequential full-parameter updates on each task without explicit preservation of prior knowledge. \textbf{Sequential LoRA} \cite{hu2022lora} serves as a parameter-efficient alternative, performing sequential adaptation with low-rank adapters. \textbf{DMT} \cite{DMT} incorporates a two-stage design: it first conducts full-parameter multi-task tuning on multi-domain SFT dataset, followed by a second stage where a small proportion of the multi-task data is replayed to mitigate forgetting.

\noindent \textbf{Evaluation Benchmarks.}
Evaluation is conducted using domain-specific benchmarks corresponding to each SFT task. For code generation, we report pass@1 on \textbf{HumanEval} \cite{chen2021evaluating}; for mathematical reasoning, we report accuracy on \textbf{GSM8K} \cite{wei2022chain}; and for medical question answering, we evaluate accuracy on the \textbf{MedQA} \cite{jin2021disease} test set. 
In our setup, each task is fine-tuned individually and sequentially, after which the model is evaluated on the aforementioned benchmarks. 

Details of benchmarks and implementation details are available in \textbf{Appendix} \ref{detail_benchmark} and \ref{implementation}. 


%% file: result.tex
\section{Results}
\subsection{Overall Performance}
We conduct extensive experiments to evaluate the effectiveness of our proposed TRACE framework in mitigating catastrophic forgetting and improving continual generalization. 
In our experiments, the sequential fine-tuning is randomly arranged in an order: math$\rightarrow$medical$\rightarrow$code. Here, TRACE-LF and TRACE-CS denote our model that performs task-specific core parameters identification using the L$_2$-Fisher importance and the cosine similarity, respectively.

As shown in Table~\ref{table1}, TRACE consistently outperforms all baselines across different model sizes and tasks. We attribute this phenomenon to our proposed strategy, which performs selective parameter activation for task-specific fine-tuning. It can effectively identify the core parameters relevant to each task and selectively activates them, thereby preventing these parameters from being overwritten during subsequent training. This mechanism substantially mitigates the catastrophic forgetting issue commonly observed in sequential fine-tuning scenarios.

Notably, TRACE achieves superior performance on the GSM8K benchmark compared to other baselines. A plausible explanation is that, in the sequential multi-task setting, the mathematical task is positioned as the first stage in the fine-tuning pipeline, making its parameter representations particularly susceptible to being overwritten during subsequent task updates. In contrast, TRACE effectively mitigates this issue by identifying and preserving task-specific core parameters, thus maintaining early task proficiency while adapting to new tasks.

Moreover, the effectiveness of TRACE is observed across diverse architectures (DeepSeek, LLaMA, Qwen) and a wide range of model sizes---from the compact 1.5B DeepSeek-R1 to the large-scale 32B Qwen3---demonstrating the scalability, robustness, and generality of our method for real-world multi-task fine-tuning scenarios.

\begin{table}[ht]
\centering
\caption{Performance of different training strategies across various LLMs. The underlined values indicate the best baseline. \textcolor{DarkGreen}{$\uparrow$} and \textcolor{DarkBlue}{$\downarrow$} represent the absolute improvement and degradation compared to the best baseline.}
\label{table1}
\resizebox{0.47\textwidth}{!}{
\begin{tabular}{cc|cccc}
\toprule
\multirow{2}{*}{LLM} & \multirow{2}{*}{Methods} & \multicolumn{4}{c}{Evaluated Benchmark $\uparrow$} \\* \cmidrule(l){3-6}
 &  & GSM8K & HumanEval & MedQA & Average \\* \midrule

\multirow{6}{*}{DeepseekR1-1.5B}
 & Joint Fine-tuning        & 48.14 & 39.00 & 27.34 & 38.16 \\
 & Sequential Fine-tuning   & 47.31 & 40.20 & \underline{27.73} & 38.41 \\
 & Sequential LoRA          & \underline{51.78} & \underline{43.40} & 26.94 & \underline{40.71} \\
 & DMT                      & 49.58 & 42.70 & 26.76 & 39.68 \\
\cmidrule(l){2-6}
 & \textbf{TRACE-LF} & 63.46\,\textcolor{DarkGreen}{\scalebox{.8}{$\uparrow$11.68}} & 43.90\,\textcolor{DarkGreen}{\scalebox{.8}{$\uparrow$0.50}} & 29.93\,\textcolor{DarkGreen}{\scalebox{.8}{$\uparrow$2.20}} & 45.76\,\textcolor{DarkGreen}{\scalebox{.8}{$\uparrow$5.05}} \\
 & \textbf{TRACE-CS} & 61.56\,\textcolor{DarkGreen}{\scalebox{.8}{$\uparrow$9.78}}  & 44.50\,\textcolor{DarkGreen}{\scalebox{.8}{$\uparrow$1.10}} & 30.40\,\textcolor{DarkGreen}{\scalebox{.8}{$\uparrow$2.67}} & 45.49\,\textcolor{DarkGreen}{\scalebox{.8}{$\uparrow$4.78}} \\*
\midrule

\multirow{6}{*}{LLaMA2-7B}
 & Joint Fine-tuning        & 29.88 & \underline{14.63} & \underline{43.05} & 29.19 \\
 & Sequential Fine-tuning   & 23.43 & 14.02 & 41.56 & 26.34 \\
 & Sequential LoRA          & 23.20 & 14.02 & 27.42 & 21.55 \\
 & DMT                      & \underline{44.12} & 12.20 & 41.71 & \underline{32.68} \\
\cmidrule(l){2-6}
 & \textbf{TRACE-LF} & 44.43\,\textcolor{DarkGreen}{\scalebox{.8}{$\uparrow$0.31}} & 15.85\,\textcolor{DarkGreen}{\scalebox{.8}{$\uparrow$1.22}} & 45.40\,\textcolor{DarkGreen}{\scalebox{.8}{$\uparrow$2.35}} & 35.23\,\textcolor{DarkGreen}{\scalebox{.8}{$\uparrow$2.55}} \\
 & \textbf{TRACE-CS} & 48.82\,\textcolor{DarkGreen}{\scalebox{.8}{$\uparrow$4.70}} & 17.07\,\textcolor{DarkGreen}{\scalebox{.8}{$\uparrow$2.44}} & 43.99\,\textcolor{DarkGreen}{\scalebox{.8}{$\uparrow$0.94}} & 36.63\,\textcolor{DarkGreen}{\scalebox{.8}{$\uparrow$3.95}} \\*
\midrule

\multirow{6}{*}{LLaMA3-8B}
 & Joint Fine-tuning        & 23.28 & 28.05 & \underline{57.66} & 36.33 \\
 & Sequential Fine-tuning   & 12.59 & 23.17 & 46.66 & 27.47 \\
 & Sequential LoRA          & 41.17 & \underline{43.29} & 54.52 & \underline{46.33} \\
 & DMT                      & \underline{49.89} & 27.44 & 55.30 & 44.21 \\
\cmidrule(l){2-6}
 & \textbf{TRACE-LF} & 58.00\,\textcolor{DarkGreen}{\scalebox{.8}{$\uparrow$8.11}}  & 57.90\,\textcolor{DarkGreen}{\scalebox{.8}{$\uparrow$14.61}} & 58.37\,\textcolor{DarkGreen}{\scalebox{.8}{$\uparrow$0.71}} & 58.09\,\textcolor{DarkGreen}{\scalebox{.8}{$\uparrow$11.76}} \\
 & \textbf{TRACE-CS} & 66.03\,\textcolor{DarkGreen}{\scalebox{.8}{$\uparrow$16.14}} & 61.60\,\textcolor{DarkGreen}{\scalebox{.8}{$\uparrow$18.31}} & 62.21\,\textcolor{DarkGreen}{\scalebox{.8}{$\uparrow$4.55}} & 63.28\,\textcolor{DarkGreen}{\scalebox{.8}{$\uparrow$16.95}} \\*
\midrule

\multirow{6}{*}{Qwen2.5-14B}
 & Joint Fine-tuning        & 37.98 & 34.76 & \underline{67.71} & 46.82 \\
 & Sequential Fine-tuning   & 26.00 & 32.93 & 65.75 & 41.56 \\
 & Sequential LoRA          & \underline{74.91} & \underline{45.73} & 66.38 & \underline{62.34} \\
 & DMT                      & 27.98 & 35.98 & 66.69 & 43.55 \\
\cmidrule(l){2-6}
 & \textbf{TRACE-LF} & 84.99\,\textcolor{DarkGreen}{\scalebox{.8}{$\uparrow$10.08}} & 73.20\,\textcolor{DarkGreen}{\scalebox{.8}{$\uparrow$27.47}} & 67.64\,\textcolor{DarkBlue}{\scalebox{.8}{$\downarrow$0.07}} & 75.28\,\textcolor{DarkGreen}{\scalebox{.8}{$\uparrow$12.94}} \\
 & \textbf{TRACE-CS} & 81.27\,\textcolor{DarkGreen}{\scalebox{.8}{$\uparrow$6.36}}  & 75.60\,\textcolor{DarkGreen}{\scalebox{.8}{$\uparrow$29.87}} & 69.29\,\textcolor{DarkGreen}{\scalebox{.8}{$\uparrow$1.58}}  & 75.39\,\textcolor{DarkGreen}{\scalebox{.8}{$\uparrow$13.05}} \\*
\midrule

\multirow{6}{*}{Qwen3-32B}
 & Joint Fine-tuning        & 57.06 & 73.80 & 73.37 & \underline{68.08} \\
 & Sequential Fine-tuning   & 50.64 & \underline{75.00} & 67.40 & 64.35 \\
 & Sequential LoRA          & \underline{62.40} & \underline{75.00} & 49.80 & 62.40 \\
 & DMT                      & 31.31 & 68.90 & \underline{75.96} & 58.72 \\
\cmidrule(l){2-6}
 & \textbf{TRACE-LF} & 69.52\,\textcolor{DarkGreen}{\scalebox{.8}{$\uparrow$7.12}}  & 90.90\,\textcolor{DarkGreen}{\scalebox{.8}{$\uparrow$15.90}} & 76.83\,\textcolor{DarkGreen}{\scalebox{.8}{$\uparrow$0.87}} & 79.08\,\textcolor{DarkGreen}{\scalebox{.8}{$\uparrow$11.00}} \\
 & \textbf{TRACE-CS} & 72.81\,\textcolor{DarkGreen}{\scalebox{.8}{$\uparrow$10.41}} & 81.10\,\textcolor{DarkGreen}{\scalebox{.8}{$\uparrow$6.10}}  & 76.28\,\textcolor{DarkGreen}{\scalebox{.8}{$\uparrow$0.32}} & 76.73\,\textcolor{DarkGreen}{\scalebox{.8}{$\uparrow$8.65}} \\*
\bottomrule
\end{tabular}
}
\end{table}

\subsection{Visual Comparison of Fine-Tuning Strategies on Task-Specific Representations}
\begin{figure*}[ht]
    \centering \includegraphics[width=0.95\textwidth]{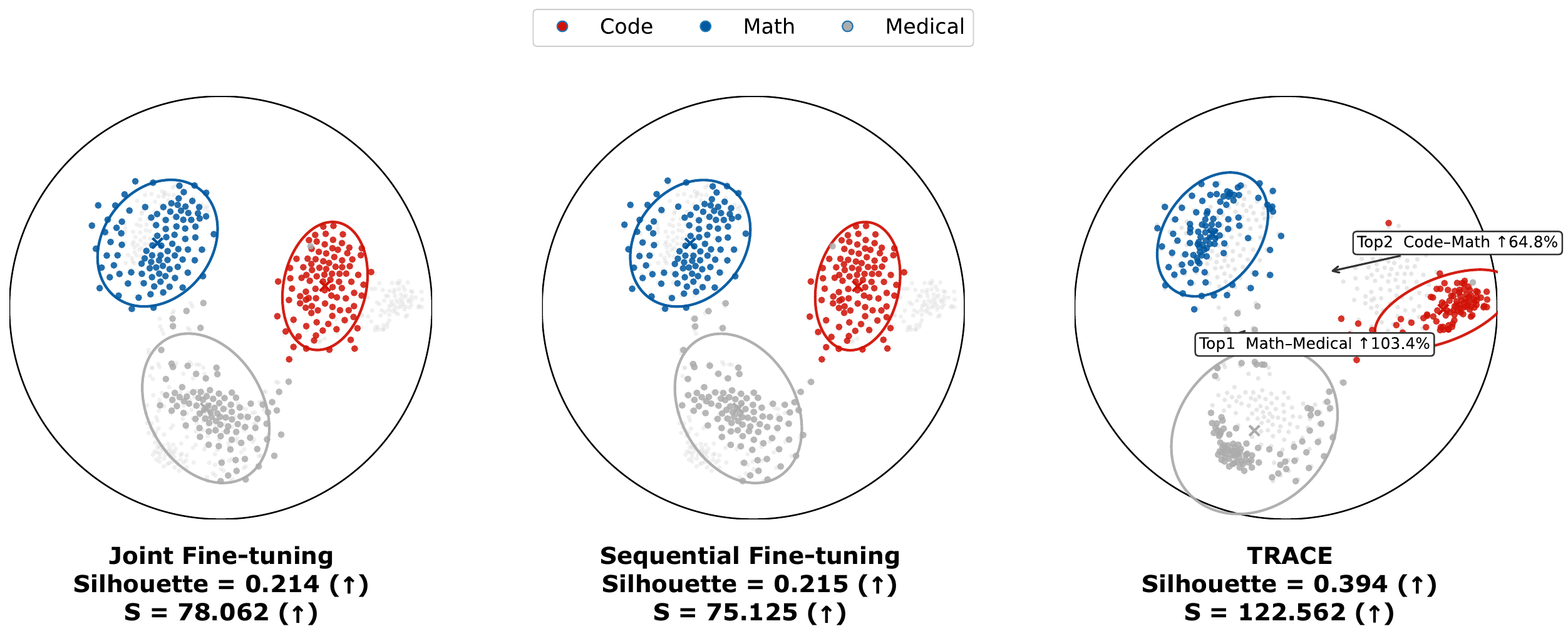}
    \caption{Visualization of Qwen2.5-14B's last-layer embeddings under three fine-tuning strategies on three domains---code generation, mathematical reasoning, and medical QA. Points are shown in a unified t-SNE space; colored dots and 95\% covariance ellipses denote clusters, with gray dots indicating the other two models for reference. Subtitles report Silhouette and the mean inter-centroid distance S (↑ better). TRACE yields the strongest separation (Silhouette = 0.394; S= 122.562), with Math–Medical and Code–Math gaps increased by 103.4\% and 64.8\% over the average of the other strategies.}
    \label{tnse}
\end{figure*}

To better understand the representational behaviors of different fine-tuning strategies, we visualize the last-layer embeddings of Qwen2.5-14B using t-SNE across three domains (Fig. \ref{tnse}). The visualizations for \textit{\textit{Joint Fine-tuning}} and\textit{ \textit{Sequential Fine-tuning}} show closely gathered and overlapping clusters, quantitatively reflected in their low Silhouette scores (0.214 and 0.215, respectively). The severe entanglement of the Math and Medical clusters in the \textit{\textit{Sequential Fine-tuning}} case provides a clear visual representation of catastrophic forgetting.
In stark contrast, our TRACE approach yields three distinct, well-separated clusters, achieving a significantly higher Silhouette score of 0.394 and a much larger inter-centroid distance (S = 122.562). This demonstrates a substantial improvement in representational separation, with the Math-Medical and Code-Math gaps increasing by 103.4\% and 64.8\% respectively, over the average of the other strategies. TRACE successfully learns robust, clearly delineated representations for each domain. Its ability to maintain this high degree of overall separation, a feat not achieved by the baseline methods where entire clusters collapse, underscores the effectiveness of its selective activation strategy.

\subsection{Task Order Sensitivity}
To explore the impact of task ordering on the performance of the TRACE framework, we conduct a sensitivity analysis by varying the fine-tuning sequence of the three tasks. As summarized in Table~\ref{table2}, we evaluate six different task permutations using the LLaMA3-8B model under TRACE.
Overall, TRACE remains robust across permutations: the averaged score over the three benchmarks is $56.54\pm3.80$ (min/max: 53.04/62.27), indicating no single ordering is required to obtain competitive multi-task performance. Moreover, all task orders consistently outperform the baseline method, demonstrating that the gains of TRACE are not tied to any particular permutation.

Importantly, every permutation yields strong results on all three benchmarks (all averages $\geq 53.04$), and multiple orders achieve similarly high overall performance (e.g., 58.09 and 58.90).
These results suggest that TRACE does not hinge on a hand-picked task sequence; instead, its core-parameter-based isolation mitigates destructive interference and preserves reasonable performance under diverse task arrangements, aligning with realistic continual settings where the task stream is dictated by data arrival.

\begin{table}[ht]
\centering
\small
\caption{Performance comparison across different task permutation orders under TRACE.}
\label{table2}
\begin{tabular}{c|cccc}
\toprule
\multirow{2}{*}{Task Order} & \multicolumn{4}{c}{Evaluated Benchmark $\uparrow$} \\* \cmidrule(l){2-5}
 & GSM8K & HumanEval & MedQA & Average \\* \midrule
Math-Medical-Code & 58.00 & 57.90 & 58.37 & 58.09 \\
Math-Code-Medical & 47.38 & 52.44 & 60.88 & 53.57 \\
Medical-Math-Code & 46.85 & 51.22 & 61.98 & 53.35 \\
Medical-Code-Math & 63.84 & 61.00 & 61.98 & 62.27 \\
Code-Math-Medical & 49.28 & 48.17 & 61.67 & 53.04 \\
Code-Medical-Math & 61.11 & 53.70 & 61.90 & 58.90 \\* \bottomrule
\end{tabular}
\end{table}

\subsection{Scalability to Long Task Sequences}
\label{sec:longseq}
A practical continual fine-tuning pipeline often involves more than a few tasks, and new tasks may be either highly heterogeneous or closely related to previously seen domains. To examine TRACE under a longer and more realistic task stream, we extend the original three-task sequence by adding two additional tasks: (\textit{i}) an in-domain math task (MATH \cite{Hendrycks2021MeasuringMP}) that is closely related to GSM8K, creating a high-overlap setting, and (\textit{ii}) a lawyer-domain task Lawyer-Instruct \cite{alignmentlab_lawyer_instruct} evaluated on LegalBench \cite{guha2023legalbench}, introducing a new domain. All methods follow the same continual fine-tuning setting.

\begin{table}[ht]
\centering
\caption{Performance comparison on the extended long task stream (LLaMA3-8B).}
\label{tab:longseq}
\resizebox{0.47\textwidth}{!}{
\begin{tabular}{c|ccccc}
\toprule
\multirow{2}{*}{Methods} & \multicolumn{5}{c}{Evaluated Benchmark $\uparrow$} \\* \cmidrule(l){2-6}
 & GSM8K & MedQA & HumanEval & LegalBench & Average \\* \midrule
Joint Fine-tuning      & 17.89 & 53.10 & 18.30 & 48.59 & 34.47 \\
Sequential Fine-tuning & 27.29 & 30.95 & 12.20 & 50.70 & 30.28 \\
Sequential LoRA        & \underline{56.63} & \underline{56.25} & \underline{47.00} & \underline{56.66} & \underline{54.66} \\
DMT                    & 32.37 & 49.73 & 14.00 & 51.42 & 36.88 \\
\cmidrule(l){1-6}
\textbf{TRACE-LF} & 66.03\,\textcolor{DarkGreen}{\scalebox{.8}{$\uparrow$9.40}}  & 58.37\,\textcolor{DarkGreen}{\scalebox{.8}{$\uparrow$2.12}} & 53.70\,\textcolor{DarkGreen}{\scalebox{.8}{$\uparrow$6.70}}  & 59.26\,\textcolor{DarkGreen}{\scalebox{.8}{$\uparrow$2.60}}  & 58.81\,\textcolor{DarkGreen}{\scalebox{.8}{$\uparrow$4.15}} \\
\textbf{TRACE-CS} & 66.72\,\textcolor{DarkGreen}{\scalebox{.8}{$\uparrow$10.09}} & 58.92\,\textcolor{DarkGreen}{\scalebox{.8}{$\uparrow$2.67}} & 61.00\,\textcolor{DarkGreen}{\scalebox{.8}{$\uparrow$14.00}} & 59.26\,\textcolor{DarkGreen}{\scalebox{.8}{$\uparrow$2.60}}  & 61.48\,\textcolor{DarkGreen}{\scalebox{.8}{$\uparrow$6.82}} \\
\bottomrule
\end{tabular}
}
\end{table}

Table~\ref{tab:longseq} shows that TRACE remains robust as the task stream grows. TRACE-CS achieves the best overall average score, outperforming the strongest baseline Sequential LoRA by +6.82 points.

Beyond the aggregate gains, this setting also clarifies how the capacity upper bound behaves in long streams. A common concern is that, as the number of tasks increases, the union of task-critical parameter sets may expand and eventually approach a large fraction of the model, diminishing the benefit of selective updates. However, such growth is largely driven by \emph{task diversity} rather than \emph{task count}. When tasks are highly related (\textit{e.g.}, GSM8K and MATH), their task-critical parameter sets are expected to exhibit substantial overlap, and the additional training acts more like rehearsal over shared subspaces than allocating new capacity. This aligns with TRACE’s design that allows parameter reuse across tasks: shared core parameters can be re-activated and refined when a related task reappears, while task-irrelevant parameters remain protected. As a result, adding an in-domain task does not degrade performance; instead, it further strengthens the corresponding capability, while maintaining strong performance on the introduced legal domain.

\subsection{Generalization and Transferability of Task-Specific Core Parameters}
TRACE's success raises a compelling scientific question: does it identify a model-specific artifact or a more fundamental, model-agnostic “functional blueprint” of a task? To investigate this, we propose and evaluate a “small-to-large” transfer paradigm. We hypothesize that core parameters identified on a smaller source model can be effectively transferred to guide the adaptation of larger, architecturally different target models. If successful, this approach would not only validate the “functional blueprint” theory but also unlock a highly resource-efficient pathway for model adaptation, decoupling the cost of analysis from the scale of deployment.

\begin{figure}[ht]
    \centering
    \includegraphics[width=0.3\textwidth]{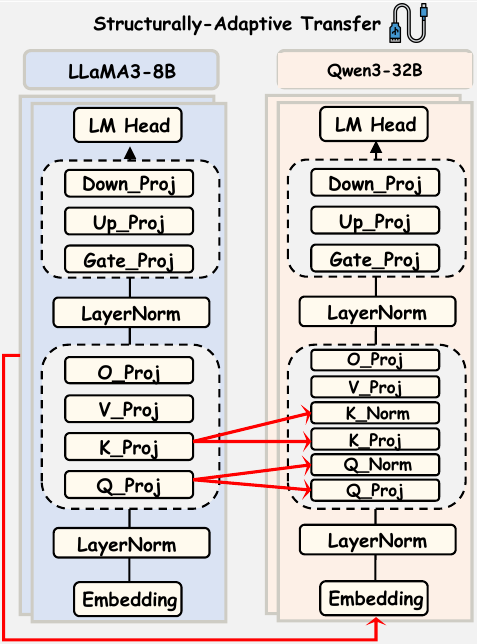}
    \caption{Illustration of our Structurally-Adaptive Transfer from a LLaMA3-8B layer to a Qwen3-32B layer.}
    \label{arc}
\end{figure}

To validate this, we introduce the \textbf{TRACE-Transfer} methodology, a two-step process designed to map core parameters across disparate architectures. The entire process is illustrated in  Fig. \ref{arc}. First, for \textit{Layer Mapping}, we identify functionally equivalent layers between the source (LLaMA3-8B) and target (Qwen series) models. This is achieved by comparing their layer activations on a small, representative probe dataset using Centered Kernel Alignment (CKA) \cite{CKA}, a robust metric adept at measuring the similarity of neural representations even when their dimensions differ. For each source layer, we select the target layer with the highest CKA similarity as its functional counterpart. Second, for \textit{Parameter Name Mapping}, we apply a set of architectural rules to translate the parameter names within the mapped layers. This rule-based approach is crucial for handling structural variations, such as the addition of \texttt{q\_norm} and \texttt{k\_norm} layers in the Qwen3-32B model, by expanding the mapping to cover the entire functional component.

The results of our transfer experiment, where core parameters for the task are identified on LLaMA3-8B and transferred to Qwen2.5-14B and Qwen3-32B, are summarized in Table~\ref{table3}. the Transfer-LF/CS methods achieve performance that is highly comparable to the standard TRACE-LF/CS methods, which necessitate a full identification process on the target model. Furthermore, the average performance of the transfer models exceeds that of all baseline methods on both the Qwen2.5-14B and Qwen3-32B benchmarks. This finding suggests that the transferred core parameters can serve as an effective foundation for multi-task adaptation, yielding stronger results than the compared baseline techniques.

\begin{table}[htbp]
\centering
\caption{Performance comparison of the transfer effectiveness across various LLMs.}
\label{table3}
\resizebox{0.47\textwidth}{!}{
\begin{tabular}{cc|cccc}
\toprule
\multirow{2}{*}{LLM} & \multirow{2}{*}{Methods} & \multicolumn{4}{c}{Evaluated Benchmark ↑} \\* \cmidrule(l){3-6}
 &  & GSM8K & HumanEval & MedQA & Average \\* \midrule
\multirow{8}{*}{Qwen2.5-14B}
 & Joint Fine-tuning   & 37.98 & 34.76 & \underline{67.71} & 46.82 \\
 & Sequential Fine-tuning   & 26.00 & 32.93 & 65.75 & 41.56 \\
 & Sequential LoRA                  & \underline{74.91} & \underline{45.73} & 66.38 & \underline{62.34}\\
 & DMT                   & 27.98 & 35.98 & 66.69 & 43.55  \\
 \cmidrule(l){2-6} 
 & \textbf{TRACE-LF}              & 84.99\,\textcolor{DarkGreen}{\scalebox{.8}{$\uparrow$10.08}} & 73.20\,\textcolor{DarkGreen}{\scalebox{.8}{$\uparrow$27.47}} & 67.64\,\textcolor{DarkBlue}{\scalebox{.8}{$\downarrow$0.07}} & 75.28\,\textcolor{DarkGreen}{\scalebox{.8}{$\uparrow$12.94}}\\
& \textbf{TRACE-CS}              & 81.27\,\textcolor{DarkGreen}{\scalebox{.8}{$\uparrow$6.36}} & 75.60\,\textcolor{DarkGreen}{\scalebox{.8}{$\uparrow$29.87}} & 69.29\,\textcolor{DarkGreen}{\scalebox{.8}{$\uparrow$1.58}} & 75.39\,\textcolor{DarkGreen}{\scalebox{.8}{$\uparrow$13.05}}\\
\cmidrule(l){2-6} 
& \textbf{Transfer-LF}           & 82.03\,\textcolor{DarkGreen}{\scalebox{.8}{$\uparrow$7.12}} & 79.90\,\textcolor{DarkGreen}{\scalebox{.8}{$\uparrow$34.17}} & 67.01\,\textcolor{DarkBlue}{\scalebox{.8}{$\downarrow$0.70}}  & 76.31\,\textcolor{DarkGreen}{\scalebox{.8}{$\uparrow$13.97}}\\
& \textbf{Transfer-CS}           & 83.40\,\textcolor{DarkGreen}{\scalebox{.8}{$\uparrow$8.49}} & 69.50\,\textcolor{DarkGreen}{\scalebox{.8}{$\uparrow$23.77}} & 65.75\,\textcolor{DarkBlue}{\scalebox{.8}{$\downarrow$1.96}} & 72.88\,\textcolor{DarkGreen}{\scalebox{.8}{$\uparrow$10.54}}\\* \midrule
\multirow{8}{*}{Qwen3-32B}
 & Joint Fine-tuning   & 57.06 & 73.80 & 73.37 & \underline{68.08}\\
 & Sequential Fine-tuning   & 50.64 & \underline{75.00} & 67.40 & 64.35\\
 & Sequential LoRA                  & \underline{62.40} & \underline{75.00} & 49.80 & 62.40 \\
 & DMT                   & 31.31 & 68.90 & \underline{75.96} & 58.72 \\
 \cmidrule(l){2-6} 
& \textbf{TRACE-LF} & 69.52\,\textcolor{DarkGreen}{\scalebox{.8}{$\uparrow$7.12}} & 90.90\,\textcolor{DarkGreen}{\scalebox{.8}{$\uparrow$15.90}} & 76.83\,\textcolor{DarkGreen}{\scalebox{.8}{$\uparrow$0.55}} & 79.08\,\textcolor{DarkGreen}{\scalebox{.8}{$\uparrow$11.00}} \\
& \textbf{TRACE-CS}         & 72.81\,\textcolor{DarkGreen}{\scalebox{.8}{$\uparrow$10.41}} & 81.10\,\textcolor{DarkGreen}{\scalebox{.8}{$\uparrow$6.10}} & 76.28\,\textcolor{DarkGreen}{\scalebox{.8}{$\uparrow$0.32}} & 76.73\,\textcolor{DarkGreen}{\scalebox{.8}{$\uparrow$8.65}}\\
\cmidrule(l){2-6} 
& \textbf{Transfer-LF}       & 68.23\,\textcolor{DarkGreen}{\scalebox{.8}{$\uparrow$5.83}} & 90.20\,\textcolor{DarkGreen}{\scalebox{.8}{$\uparrow$7.12}} & 77.77\,\textcolor{DarkGreen}{\scalebox{.8}{$\uparrow$1.49}} & 76.31\,\textcolor{DarkGreen}{\scalebox{.8}{$\uparrow$8.23}}\\
& \textbf{Transfer-CS}       & 68.99\,\textcolor{DarkGreen}{\scalebox{.8}{$\uparrow$6.59}} & 88.40\,\textcolor{DarkGreen}{\scalebox{.8}{$\uparrow$13.40}} & 75.26\,\textcolor{DarkBlue}{\scalebox{.8}{$\downarrow$0.70}}  & 73.23\,\textcolor{DarkGreen}{\scalebox{.8}{$\uparrow$5.15}}\\* \bottomrule
\end{tabular}
}
\end{table}

\subsection{Hyperparameter Sensitivity}
\subsubsection{Sensitivity to Core Parameter Retention Ratio}
\label{sec:r_sens}
Recall that we update only a small subset of parameters selected by the importance
ranking, and denote the \emph{filtering ratio} $r$ as the percentage of parameters kept
trainable, while the remaining parameters are frozen.
To verify that our default choice is not arbitrary, we sweep
$r\in\{1,5,10,15,20,30,40,50,60,70,80,90,100\}$, keeping all other settings fixed.

\begin{figure}[ht]
    \centering \includegraphics[width=0.47\textwidth]{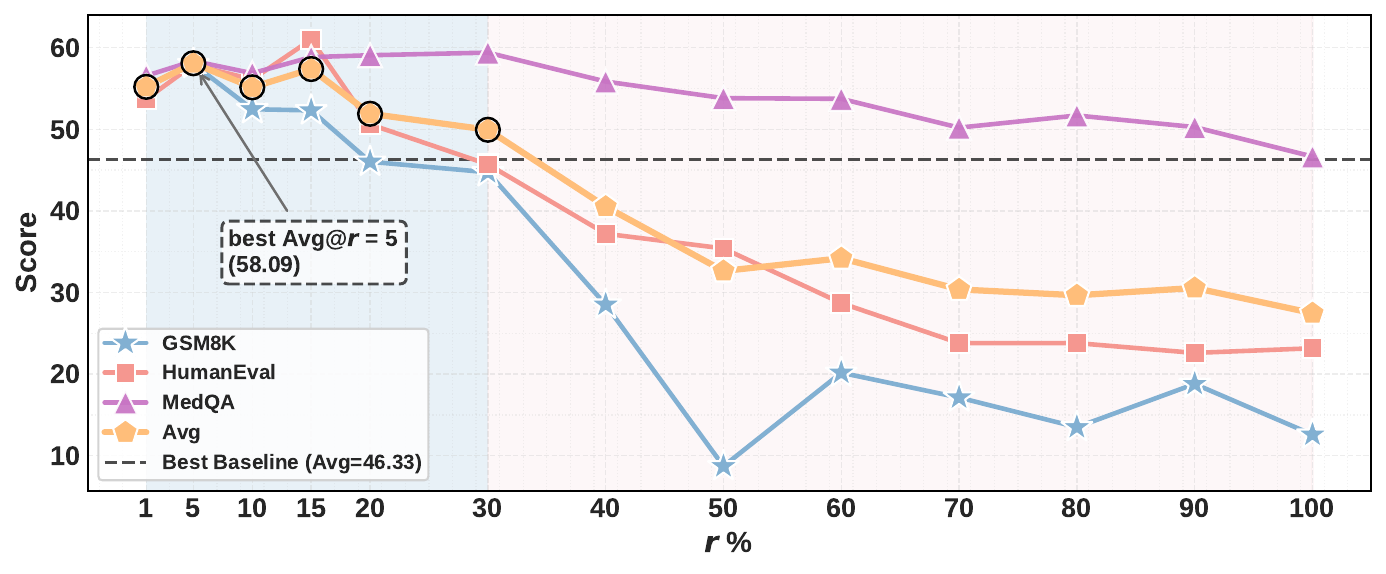}
    \caption{Scores with varying ratios of top $r$\% parameters.}
    \label{top}
\end{figure}

Fig.~\ref{top} reports the resulting performance trends, and additionally overlays the
best baseline average score as a dashed line for direct comparison.
We observe that the average score peaks at $r=5$ and remains consistently above the
baseline within a small retention range ($r\in[1,30]$), indicating that the method is not
overly sensitive in the practical regime and that a fixed default is reasonable.
In contrast, once $r$ becomes large, performance degrades sharply, and the average falls below the baseline when $r\ge 40$, suggesting that updating
a substantial number of low-importance parameters introduces strong interference.
Therefore, we set $r=5$ by default, and report the full sweep in
\textbf{Appendix}~\ref{r_scan}.

\subsubsection{Sensitivity to Warm-Start Epochs}
Our warm-start stage is designed to \emph{probe} task-sensitive parameters by inducing a small amount of task-specific updates, rather than to optimize the task loss to convergence. 
A natural question is how the warm-start length affects the identified core parameters and the final sequential adaptation performance. 
To this end, we vary the warm-start epochs $E_{\mathrm{ws}}\in\{1,2,3\}$ for the three task on LLaMA3-8B while keeping all other settings fixed, re-identify the core set under each setting, and evaluate the resulting pipeline on GSM8K, HumanEval, and MedQA.

\begin{figure}[ht]
    \centering \includegraphics[width=0.49\textwidth]{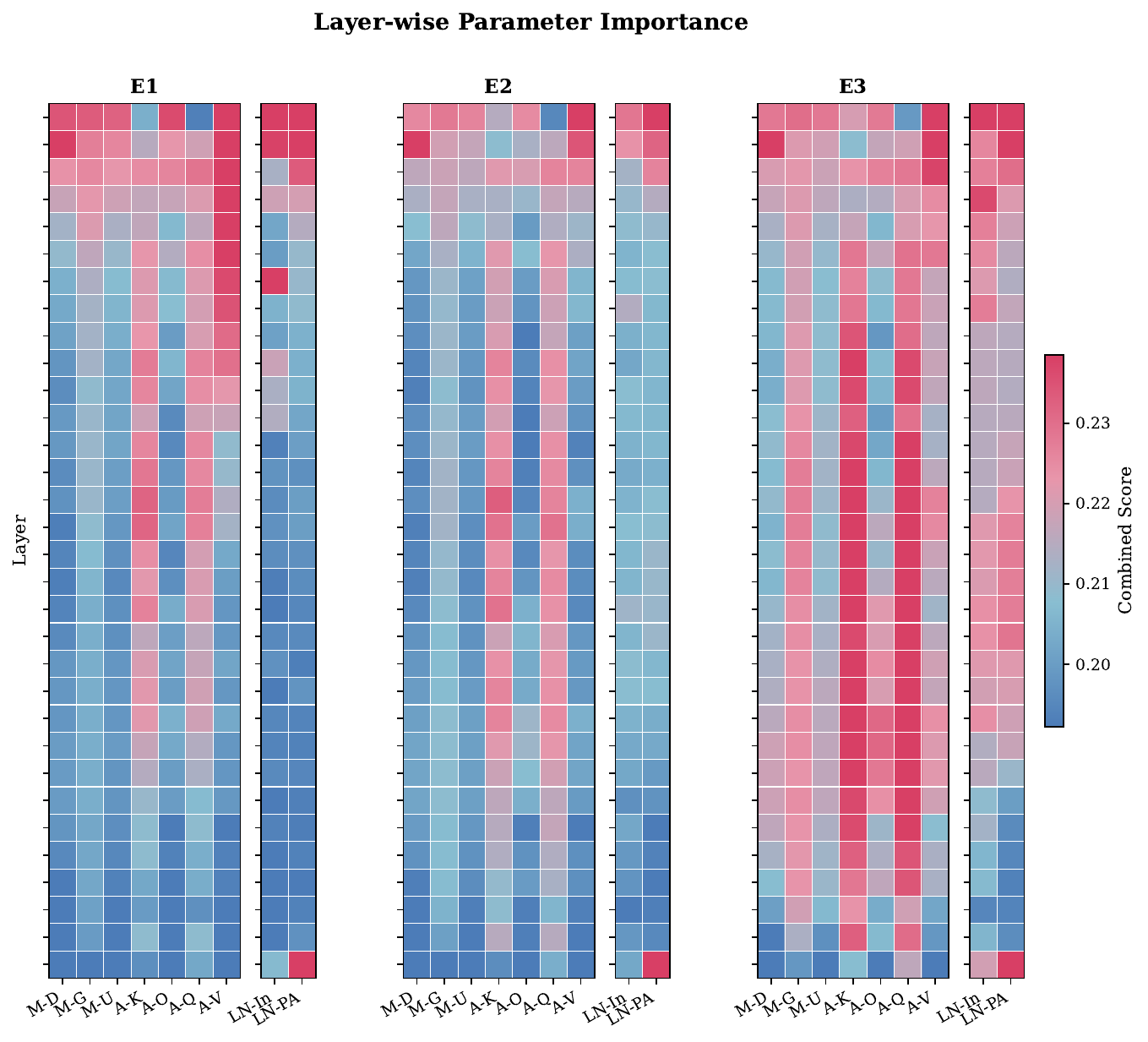}
    \caption{Layer-wise parameter importance across three training epochs. Each column shows one epoch with projection parameters (left) and LayerNorm parameters (right). Module abbreviations: A-Q/K/V/O       
  (attention q/k/v/o projections), M-D/G/U (MLP down/gate/up), LN-In/PA (input/post-attention layernorm). Color intensity is the combined importance scores.}
    \label{warmstart}
\end{figure}

Fig.~\ref{warmstart} visualizes the code task's layer-wise importance distribution (projection modules on the left and LayerNorm modules on the right) computed after each warm-start length. 
With $E_{\mathrm{ws}}=1$, the importance pattern remains relatively \emph{structured and sparse}: a subset of modules/layers exhibit clearly higher scores, yielding a core set that is more discriminative for the probed task.
As $E_{\mathrm{ws}}$ increases to 2 and 3, the importance distribution becomes progressively more \emph{diffuse} across layers and modules.
This ``spreading'' effect indicates that longer warm-start tends to activate broader parameter updates, making many parameters appear important and thereby reducing the separability between truly task-critical parameters and generally plastic parameters.
In other words, overly long warm-start can dilute the identification signal by mixing early task-specific sensitivity with later-stage, more global adaptation dynamics.

The quantitative results in Table~\ref{tab:warmstart_epochs} confirm this trend for both strategies.
For TRACE-LF, $E_{\mathrm{ws}}=1$ yields the best average. TRACE-CS shows the same pattern: the best average is achieved at $E_{\mathrm{ws}}=1$ and decreases as $E_{\mathrm{ws}}$ increases.
This matches Fig.~\ref{warmstart}: as importance becomes more saturated across layers/modules (notably at $E_{\mathrm{ws}}=3$), the core set becomes less task-discriminative and more interference-prone, hurting generalization.
Heatmaps for the \textit{math} and \textit{medical} tasks are provided in \textbf{Appendix}~\ref{detail_warm-start}.

\begin{table}[ht]
\centering
\small
\caption{Effect of warm-start epochs $E_{\mathrm{ws}}$ on the benchmarks (LLaMA3-8B).}
\label{tab:warmstart_epochs}
\begin{tabular}{c|c|ccc|c}
\toprule
\multirow{2}{*}{Methods} & \multirow{2}{*}{$E_{\mathrm{ws}}$} & \multicolumn{3}{c|}{Evaluated Benchmark $\uparrow$} & \multirow{2}{*}{Average $\uparrow$} \\* \cmidrule(l){3-5}
 &  & GSM8K & HumanEval & MedQA &  \\* \midrule
\multirow{3}{*}{TRACE-LF} 
& 1 & 58.00 & 57.90 & 58.37 & \textbf{58.09} \\
& 2 & 56.86 & 51.80 & 59.15 & 55.94 \\
& 3 & 49.96 & 49.40 & 59.31 & 52.89 \\ \midrule
\multirow{3}{*}{TRACE-CS} 
& 1 & 66.03 & 61.60 & 62.21 & \textbf{63.28} \\
& 2 & 62.17 & 61.00 & 59.15 & 60.77 \\
& 3 & 59.14 & 56.10 & 59.94 & 58.71 \\* \bottomrule
\end{tabular}
\end{table}

\subsubsection{Sensitivity to $\alpha$ (\text{L}$_2$--Fisher Trade-off)}
\label{sec:alpha_sens}
\begin{figure}[ht]
    \centering \includegraphics[width=0.47\textwidth]{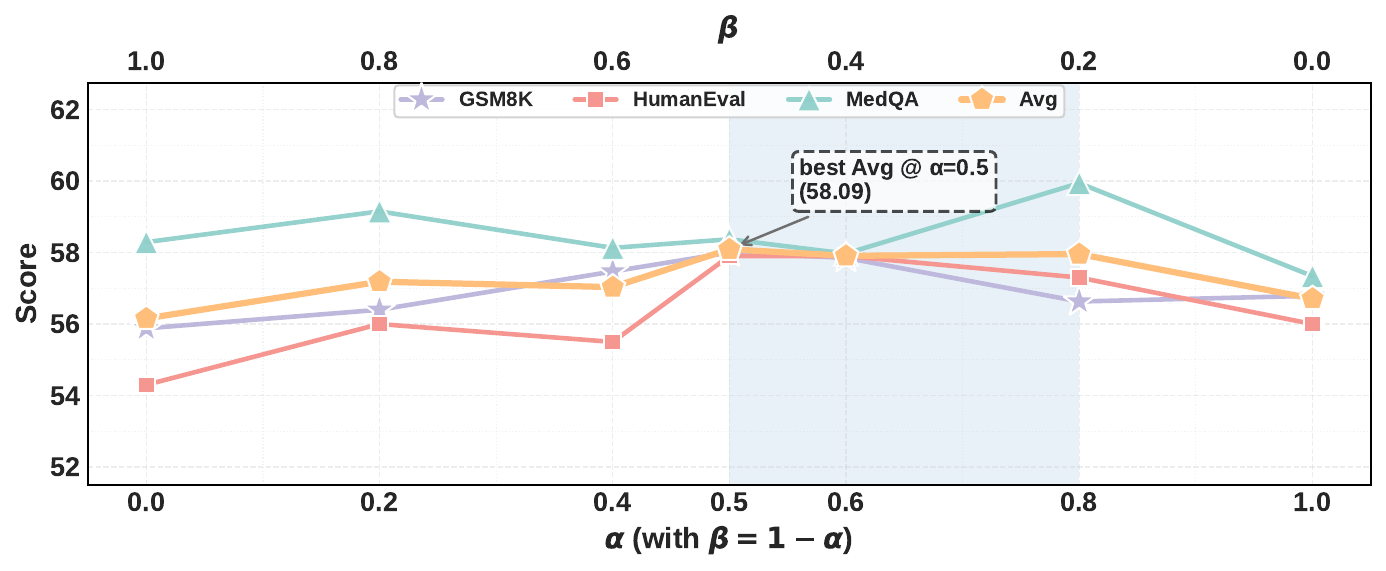}
    \caption{Sensitivity to the trade-off coefficient $\alpha$ (with $\beta=1-\alpha$) on GSM8K,
HumanEval, and MedQA.}
    \label{fig:alpha_sensitivity}
\end{figure}

Recall that we rank parameters by the combined importance score 
$s_i = \alpha \cdot \text{L}_{2,i} + (1-\alpha) \cdot \text{Fisher}_i$.
To verify that our default choice is not arbitrary, we sweep
$\alpha\in\{0,0.2,0.4,0.5,0.6,0.8,1.0\}$ (with $\beta=1-\alpha$), while keeping all other
settings fixed.
Fig.~\ref{fig:alpha_sensitivity} shows that the average score peaks at $\alpha=0.5$ and
remains stable for $\alpha\in[0.5,0.8]$, indicating that the method is not overly sensitive
to this hyperparameter. Notably, using only Fisher ($\alpha=0$) or only L$_2$  ($\alpha=1$)
yields worse average performance, suggesting that the two signals provide complementary
information. Therefore, we use $\alpha=\beta=0.5$ in all experiments; full results are reported in \textbf{Appendix}~\ref{alpha_scan}.

%% file: conclusion.tex
\section{Conclusion}
In this paper, we propose an effective framework, named TRACE, to alleviate catastrophic forgetting in continual fine-tuning of LLMs. TRACE identifies and activates only a small subset of task-specific core parameters at each stage, improving continual adaptation performance while reducing computational cost. Specifically, we present two strategies to select core parameters via adaptation-aware probing: TRACE-LF is \emph{task-local}: it estimates per-task parameter salience by combining update magnitude (L$_2$ ) and loss sensitivity (Fisher), requiring no cross-task comparisons. In contrast, TRACE-CS is \emph{task-contrastive}: it measures parameter specificity by comparing update directions across tasks (cosine similarity), explicitly favoring parameters whose updates are unique to the current task. This independence makes the framework robust to different task regimes. 

%% file: appendix.tex
\section{Appendix}

\subsection{Details of SFT Datasets}
\label{detail_sft}
\noindent
\textbf{GSM8K RFT} \cite{yuan2023scaling} extends the original GSM8K benchmark \cite{cobbe2021training} by incorporating multiple reasoning trajectories via a rejection sampling procedure. This augmentation yields a training set comprising 7.5K questions paired with 110K annotated responses. The dataset has been made publicly available to support further research in mathematical reasoning.

\noindent
\textbf{Code Alpaca} \cite{codealpaca} is a code generation dataset designed to train instruction-following LLaMA models. It is fully derived from Stanford Alpaca and includes 20K instances tailored for fine-tuning. The dataset has been publicly released for open access and research use.

\noindent
\textbf{MedQA} \cite{jin2021disease} is a free-form multiple-choice question answering dataset focused on medical problem solving, constructed from professional medical board examinations. It supports three languages---English, Simplified Chinese, and Traditional Chinese---with 12,723, 34,251, and 14,123 questions, respectively. In addition to the question set, a large-scale corpus extracted from medical textbooks is provided to facilitate knowledge-intensive reading comprehension. The MedQA dataset has been open-sourced for academic use.

\subsection{Details of Foundation Models}
In this section, we provide descriptions of the foundation language models used in our experiments to better contextualize the position of our proposed method.
\label{detail_model}

\noindent
\textbf{DeepSeek-R1} \cite{guo2025deepseek}:
This model is a distilled version of the DeepSeek R1 reasoning model, derived from Qwen2.5-Math-1.5B. It is optimized for mathematical problem solving and demonstrates strong performance on reasoning benchmarks. Notably, it achieves 83.9\% Pass@1 on MATH500 and 28.9\% Pass@1 on AIME 2024, outperforming several much larger models including GPT-4o and Claude 3.5, despite having only 1.5 billion parameters. Its compact size and high efficiency make it suitable for resource-constrained environments and reasoning-intensive tasks.

\noindent
\textbf{LLaMA2-7B} \cite{touvron2023llama}:
LLaMA2 is an open-source large language model series released by Meta AI, featuring improved training datasets and optimization techniques over the original LLaMA. The 7B variant, used in our experiments, supports a wide range of tasks such as text generation, summarization, and instruction following. LLaMA2 places emphasis on safety and bias mitigation, making it well-suited for both academic and industrial research.

\noindent
\textbf{LLaMA3-8B} \cite{dubey2024llama}:
LLaMA3 represents the latest generation in Meta’s LLaMA family, incorporating architectural enhancements, improved tokenization, and extended training on high-quality datasets. The 8B model delivers notable improvements in reasoning and generalization compared to LLaMA2, and it serves as a strong baseline for multi-task supervised fine-tuning in our study.

\noindent
\textbf{Qwen2.5-14B} \cite{qwen2.5}:
Qwen2.5 is a bilingual (Chinese-English) open-source LLM series developed by Alibaba DAMO Academy. The 14B model is instruction-tuned and pre-trained on a diverse mixture of natural language and code data. It adopts group-query attention and supports long-context modeling. Qwen2.5-14B achieves competitive performance in both zero-shot and supervised settings across a variety of benchmarks.

\noindent
\textbf{Qwen3-32B} \cite{yang2025qwen3}:
Qwen3 is the latest open-source LLM series released by Alibaba in April 2025, covering dense and Mixture-of-Experts (MoE) models ranging from 0.5B to 235B parameters. The 32B dense model employed in our experiments supports multi-lingual, reasoning, and domain-specific tasks. It introduces a hybrid inference paradigm with “thinking” and “non-thinking” modes, adaptive reasoning budget control, and exhibits state-of-the-art performance among open-source models on multiple benchmarks.

\subsection{Details of Evaluation Benchmarks}
\label{detail_benchmark}

We evaluate model performance across three distinct benchmarks, each targeting a specific domain: mathematical reasoning, code generation and medical question answering. The selected benchmarks include GSM8K, HumanEval and MedQA.

\noindent
\textbf{GSM8K} \cite{yuan2023scaling} is a widely adopted benchmark for evaluating arithmetic reasoning abilities. It comprises grade-school-level math word problems that require multi-step numerical reasoning. Our evaluation follows the standard protocol on the test set, using 8-shot chain-of-thought prompting to elicit detailed intermediate steps. This setup emphasizes a model’s ability to reason through complex arithmetic rather than relying on surface-level pattern recognition.

\noindent
\textbf{HumanEval} \cite{chen2021evaluating} is a code generation benchmark introduced to assess functional correctness in Python programming tasks. The dataset includes 164 problems, each specified with a function signature and an accompanying docstring. Model outputs are evaluated based on their ability to pass unit tests that verify semantic equivalence with reference implementations. We adopt the official evaluation script for consistent and reproducible measurement of functional correctness.

\noindent
\textbf{MedQA} \cite{jin2021disease} is a multiple-choice question answering dataset constructed from real-world medical licensing examinations. It targets clinical knowledge and diagnostic reasoning across a wide range of medical subfields. The dataset is divided into a training set and a test set, enabling both supervised fine-tuning and evaluation. In our study, we exclusively use the English-language test set for evaluation purposes, measuring the model’s ability to generalize to medical questions without additional in-domain supervision. This benchmark poses significant challenges due to its domain specificity, factual rigor, and the need for multi-hop reasoning.

\subsection{Implementation Details}
\label{implementation}
We conduct extensive fine-tuning experiments across three classic SFT datasets---Code Alpaca, GSM8K-RFT, and MedQA---using five LLM backbones of varying scales: DeepSeekR1-1.5B, LLaMA2-7B, LLaMA3-8B, Qwen2.5-14B, and Qwen3-32B. For all full-parameter fine-tuning experiments, we adopt the AdamW optimizer \cite{loshchilov2018decoupled} with a cosine learning rate schedule, a maximum sequence length of 1,024, and BFloat16 mixed precision. All training is performed for 3 epochs with a global batch size of 256. Full-parameter training is implemented with DeepSpeed ZeRO Stage 3~\cite{rajbhandari2020zero} to ensure memory and computation efficiency in large-scale settings. For LoRA-sequential, we set the rank $r=8$, scaling factor $\alpha=16$, and insert adapters into all Transformer layers. The learning rate for LoRA tuning is $2.0 \times 10^{-5}$. All experiments are conducted on 8 NVIDIA A800-SXM GPUs with 80GB memory each.

\subsection{Details of Sensitivity to Filtering Ratio}
\label{r_scan}

Table~\ref{tab:r_scan} reports the full results of sweeping the filtering ratio
$r\in\{1,5,10,15,20,30,40,50,60,70,80,90,100\}$.

\begin{table}[ht]
\centering
\caption{Effect of the filtering ratio $r$ (percentage of trainable parameters) on downstream benchmarks.}
\label{tab:r_scan}
\begin{tabular}{c|ccc|c}
\toprule
\multirow{2}{*}{$r$} & \multicolumn{3}{c|}{Evaluated Benchmark $\uparrow$} & \multirow{2}{*}{Average $\uparrow$} \\* \cmidrule(l){2-4}
 & GSM8K & HumanEval & MedQA &  \\* \midrule
Best baseline & 41.17 & 43.29 & 54.52 & 46.33 \\
\midrule
1\%   & 55.34 & 53.70 & 56.56 & 55.20 \\
5\%   & \textbf{58.00} & 57.90 & 58.37 & \textbf{58.09} \\
10\%  & 52.46 & 56.10 & 56.87 & 55.14 \\
15\%  & 52.31 & \textbf{61.00} & 58.84 & 57.38 \\
20\%  & 46.02 & 50.60 & 59.07 & 51.90 \\
30\%  & 44.73 & 45.70 & \textbf{59.39} & 49.94 \\
40\%  & 28.51 & 37.20 & 55.85 & 40.52 \\
50\%  &  8.72 & 35.40 & 53.81 & 32.64 \\
60\%  & 20.17 & 28.70 & 53.73 & 34.20 \\
70\%  & 17.13 & 23.80 & 50.20 & 30.38 \\
80\%  & 13.50 & 23.80 & 51.69 & 29.66 \\
90\%  & 18.80 & 22.60 & 50.27 & 30.56 \\
100\% & 12.59 & 23.17 & 46.66 & 27.47 \\
\bottomrule
\end{tabular}
\end{table}

\subsection{Details of Sensitivity to Warm-Start Epochs}
\label{detail_warm-start}

For completeness, we report the layer-wise parameter-importance heatmaps for the \textit{math} and \textit{medical} tasks under different warm-start epochs $E_{\mathrm{ws}}\in\{1,2,3\}$.
The visualization protocol is identical to that used in the main paper (Fig.~\ref{warmstart}): each column corresponds to one warm-start length (E1/E2/E3), with projection parameters shown on the left panel and LayerNorm parameters shown on the right panel.
Module abbreviations follow the same convention: A-Q/K/V/O denote attention q/k/v/o projections, M-D/G/U denote MLP down/gate/up projections, and LN-In/PA denote input/post-attention layernorm.

\begin{figure}[ht]
    \centering
    \includegraphics[width=0.48\textwidth]{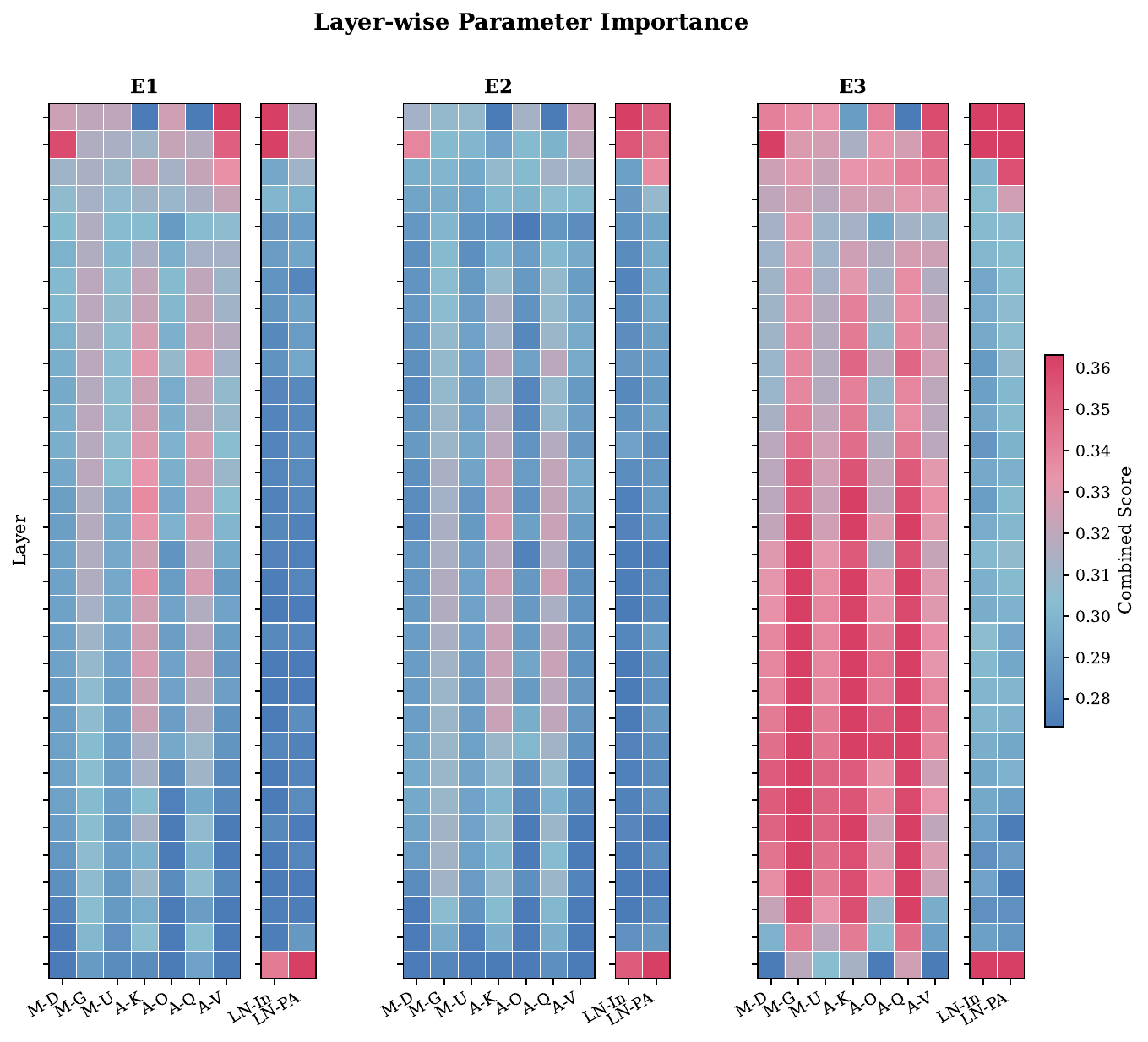}
    \caption{Layer-wise parameter importance across warm-start epochs on the \textit{math} task. Increasing $E_{\mathrm{ws}}$ makes the importance distribution less sparse and more widespread across layers/modules, indicating that longer warm-start tends to dilute the task-discriminative signal for core-parameter identification.}
    \label{fig:warmstart_math}
\end{figure}

\begin{figure}[ht]
    \centering
    \includegraphics[width=0.48\textwidth]{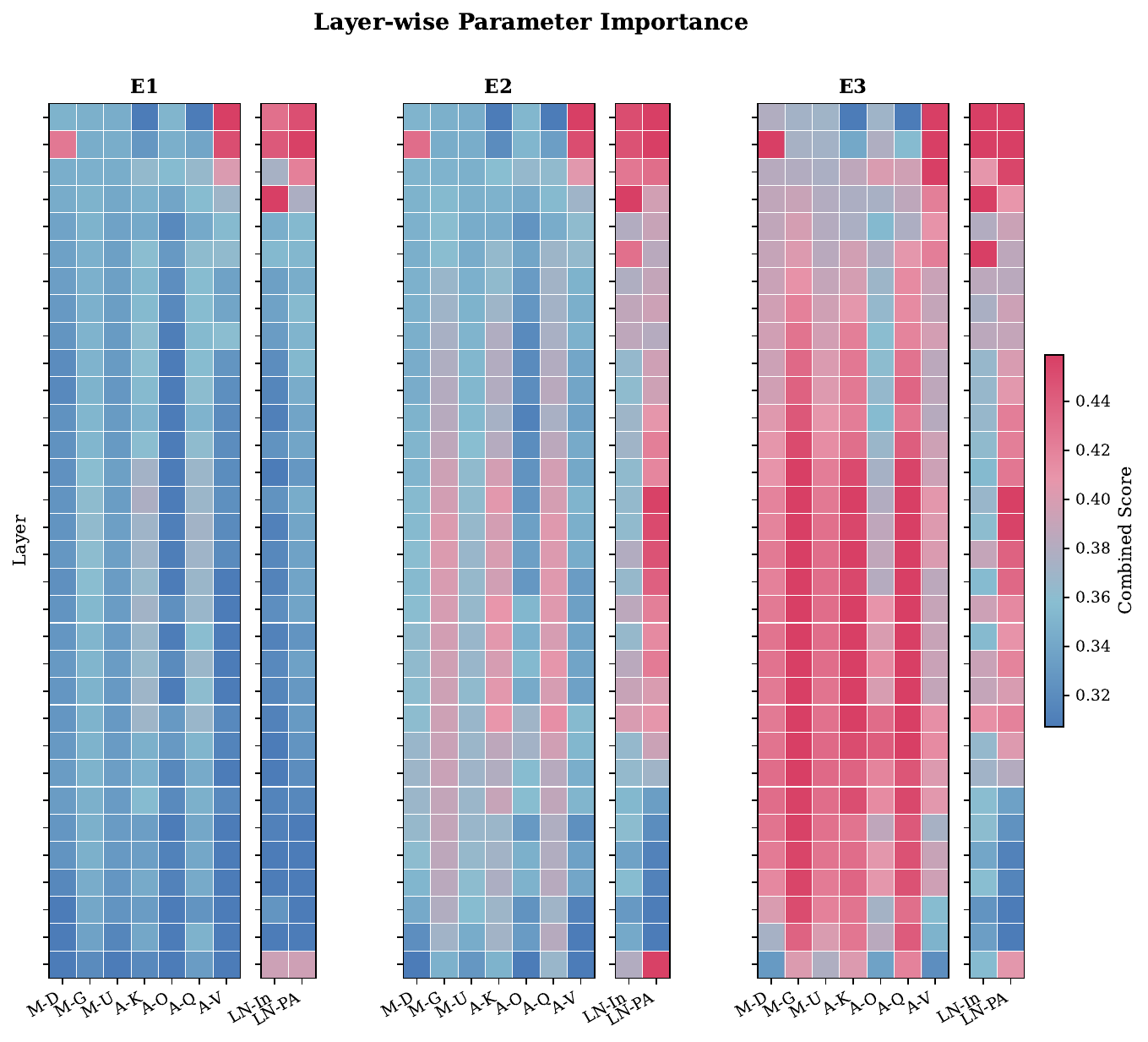}
    \caption{Layer-wise parameter importance across warm-start epochs on the \textit{medical} task. The qualitative trend is consistent with the \textit{math} and \textit{code} tasks: a short warm-start yields a more structured importance pattern, whereas longer warm-start (E3) results in a more diffuse distribution across layers/modules.}
    \label{fig:warmstart_medical}
\end{figure}

\noindent\textbf{Discussion.}
Across both tasks, we observe a consistent qualitative trend with the main-paper findings on the \textit{code} task: $E_{\mathrm{ws}}=1$ produces a relatively structured (more discriminative) importance pattern, while longer warm-start epochs progressively spread high-importance scores across broader layers and modules.
This supports our design choice of using a single-epoch warm-start as a lightweight yet effective probe for identifying core parameters, while keeping the identification overhead minimal.

\subsection{Details of Sensitivity to $\alpha$, $\beta$}
\label{alpha_scan}

Table~\ref{tab:alpha_scan} reports the full results of sweeping
$\alpha\in\{0,0.2,0.4,0.5,0.6,\\ 0.8,1.0\}$ with $\beta=1-\alpha$.

\begin{table}[ht]
\centering
\caption{Effect of the trade-off coefficient $\alpha$ (with $\beta=1-\alpha$) on downstream benchmarks.}
\label{tab:alpha_scan}
\begin{tabular}{c|ccc|c}
\toprule
\multirow{2}{*}{$\alpha$} & \multicolumn{3}{c|}{Evaluated Benchmark $\uparrow$} & \multirow{2}{*}{Average $\uparrow$} \\* \cmidrule(l){2-4}
 & GSM8K & HumanEval & MedQA &  \\* \midrule
0.0 & 55.88 & 54.30 & 58.29 & 56.16 \\
0.2 & 56.40 & 56.00 & 59.15 & 57.18 \\
0.4 & 57.47 & 55.50 & 58.13 & 57.03 \\
0.5 & \textbf{58.00} & \textbf{57.90} & 58.37 & \textbf{58.09} \\
0.6 & 57.85 & \textbf{57.90} & 57.97 & 57.91 \\
0.8 & 56.63 & 57.30 & \textbf{59.94} & 57.96 \\
1.0 & 56.79 & 56.00 & 57.34 & 56.71 \\* \bottomrule
\end{tabular}
\end{table}
